\documentclass{article}
\usepackage{amsfonts}
\usepackage{spconf,amsmath,graphicx,epsfig}
\usepackage{amsmath, amsthm, amssymb, epsfig}
\usepackage{xcolor}
\usepackage{hyperref}
\title{Classical-to-Quantum Transfer Learning for Spoken Command Recognition Based on Quantum Neural Networks}
\name{Jun Qi$^{1}$,  Javier Tejedor$^{2}$ }
\address{$^1$Electrical and Computer Engineering, Georgia Institute of Technology, Atlanta, GA, USA \\
$^2$ Escuela Politecnica Superior, Universidad San Pablo-CEU, CEU Universities, Madrid, Spain}

\begin{document}
\ninept
\maketitle
\begin{abstract}
This work investigates an extension of transfer learning applied in machine learning algorithms to the emerging hybrid end-to-end quantum neural network (QNN) for spoken command recognition (SCR). Our QNN-based SCR system is composed of classical and quantum components: (1) the classical part mainly relies on a 1D convolutional neural network (CNN) to extract speech features; (2) the quantum part is built upon the variational quantum circuit with a few learnable parameters. Since it is inefficient to train the hybrid end-to-end QNN from scratch on a noisy intermediate-scale quantum (NISQ) device, we put forth a hybrid transfer learning algorithm that allows a pre-trained classical network to be transferred to the classical part of the hybrid QNN model. The pre-trained classical network is further modified and augmented through jointly fine-tuning with a variational quantum circuit (VQC). The hybrid transfer learning methodology is particularly attractive for the task of QNN-based SCR because low-dimensional classical features are expected to be encoded into quantum states. We assess the hybrid transfer learning algorithm applied to the hybrid classical-quantum QNN for SCR on the Google speech command dataset, and our classical simulation results suggest that the hybrid transfer learning can boost our baseline performance on the SCR task. 
\end{abstract}
\section{Introduction}

The state-of-the-art baseline system of spoken command recognition (SCR)~\cite{de2018neural, mcmahan2018listening, bae2018end} is built upon the advancement of deep learning (DL) technology~\cite{goodfellow2016deep}. The DL technologies highly rely on two important aspects: (1) powerful computational resources arisen from the graphical processing unit (GPU); (2) the availability of access to a large amount of labelled or unlabelled training data~\cite{qi2019theory, qi2020analyzing}. Despite the rapid empirical progress of the SCR systems, deep learning models are becoming computational expensive and, now that Moore's law is faltering~\cite{theis2017end}, it is necessary to contemplate a future technology to further deal with a huge amount of speech data. However, new exciting possibilities are opening up due to the imminent advent of quantum computing devices that directly exploit the laws of quantum mechanics to evade the technological limits of classical computation~\cite{qi2021qtn}. 

The exploitation of quantum computing machines to carry out quantum machine learning (QML)~\cite{biamonte2017quantum} is still in its initial exploratory stage. In particular, a quantum neural network (QNN)~\cite{mcclean2018barren}, which is capable of carrying out a universal quantum computation, attracts much attention in the domain of QML because it can be seen as a quantum analog of the classical deep neural network (DNN). Hence, the QNN model allows the algorithm of back-propagation to train the model parameters of the QNN. However, in the age of noisy intermediate-scale quantum (NISQ) devices~\cite{preskill2018quantum}, it is necessary to simulate quantum experiments with noisy quantum circuits that may degrade the baseline performance. A compromised QNN, namely variational quantum circuit (VQC), has been proposed to overcome the influence of quantum noise in the QNN. The advantages of the VQC in the QNN arise from many aspects: (1) a VQC is a quantum circuit with adjustable parameters that are optimized according to a predefined metric; (2) VQC is flexibly placed on a NISQ machine and can be resilient to the quantum noisy effects on quantum circuits. 

Prominent examples in the hybrid QNN for machine learning tasks include quantum reinforcement learning~\cite{chen2020variational}, quantum image processing~\cite{cong2019quantum}, and quantum circuit learning (QCL)~\cite{mitarai2018quantum}, where the VQC model plays an important role as a quantum component. Particularly as for the application of QNNs for SCR, our pioneering work~\cite{yang2021decentralizing} investigates the use of quantum convolutional neural networks to extract quantum speech features as the input to state-of-the-art classical deep neural networks. However, the work~\cite{yang2021decentralizing} does not employ the QNN as an end-to-end SCR model, and it is still unknown about the performance of QNNs for SCR. Hence, 
in this work, we attempt to build a hybrid classical-quantum QNN model for SCR. In particular, we make use of a hybrid transfer learning approach to speed up the end-to-end training framework based on the QNN model.  

Classical transfer learning~\cite{pan2009survey, tan2018survey} is a typical example of machine learning that has been originally inspired by biological intelligence, and it originates from the knowledge acquired in a specific context which can be transferred to a different area. For example, when we learn a foreign language, we always make use of our previous linguistic knowledge to speed up the learning rate~\cite{wehrli2009deep}. This general idea has been successfully applied to artificial intelligence and machine learning domains. It has been shown that in many situations, instead of training a full network from scratch, it should be more efficient to start from a pre-trained network so that only some model components need further fine-tuning for a particular task of interest. The transfer learning methodology is very related to the training of a hybrid classical-quantum QNN, where the classical component can be pre-trained in another generic model and directly transferred to the hybrid QNN. 

This work aims to investigate a hybrid classical-to-quantum transfer learning paradigm in the context of a QNN-based SCR system. Our baseline system, namely CNN-DNN, consists of two core components: (1) convolutional neural network (CNN) for speech feature extraction; (2) DNN for recognizing commands. In this work, we first build a hybrid classical-quantum model, namely CNN-QNN, where the CNN is still kept for feature extraction and the DNN is replaced by the QNN model. In doing so, a classical CNN-DNN becomes a hybrid classical-quantum CNN-QNN. Although the hybrid model follows an end-to-end learning pipeline, the frequent data conversion between classical and quantum states greatly slows down the computing efficiency in the training stage. Moreover, the current NISQ device admits only a small number of qubits for the QNN so that the representation power of the CNN-QNN may become quite limited. Thus, a new training strategy needs to be proposed. Here, we put forth a novel hybrid transfer learning strategy. In more detail, given a well-trained classical CNN-DNN, the CNN model is taken to initialize the CNN component of the CNN-QNN and several additional training steps are conducted to fine-tune the parameters of the QNN. 

The rest of the paper is organized as follows: Section~\ref{sec2} introduces some notations. Section~\ref{sec3} introduces the architecture of the VQC-based QNN. Section~\ref{sec4} presents the framework of our SCR system. Section~\ref{sec5} shows the hybrid transfer learning algorithm. Our simulation results will be reported in Section~\ref{sec6}, and the paper is concluded in Section~\ref{sec7}.

\section{Notations for Quantum Computing}
\label{sec2}
We denote $\mathbb{R}^{d}$ as the d-dimensional real coordinate space. Given a vector $\textbf{v} = [v_{1}, v_{2}, ..., v_{d}]^{T} \in \mathbb{R}^{d}$, a $d$-qubit quantum state $|\textbf{v} \rangle = \otimes_{i=1}^{d} |v_{i}\rangle =  |v_{1}\rangle \otimes |v_{2} \rangle \otimes \cdot\cdot\cdot \otimes |v_{d} \rangle$ is a quantum state associated with a $2^{d}$-dimensional vector in a Hilbert space, where for a scalar $v_{i}$, the quantum state $|v_{i} \rangle$ can be written as Eq. (\ref{eq:1}).
\begin{equation}
\label{eq:1}
|v_{i} \rangle = \cos v_{i} |0\rangle + \sin v_{i} |1\rangle = \left[\begin{matrix} \cos v_{i} \\ \sin v_{i} \end{matrix}\right]. 
\end{equation}

\section{Variational Quantum Circuit-Based Quantum Neural Network}
\label{sec3}
Figure~\ref{fig:fig1} exhibits the architecture of the QNN including three components: (1) quantum encoding; (2) VQC; (3) measurement. The introduction of each component is shown as follows:
\begin{figure}[ht]
\centerline{\epsfig{figure=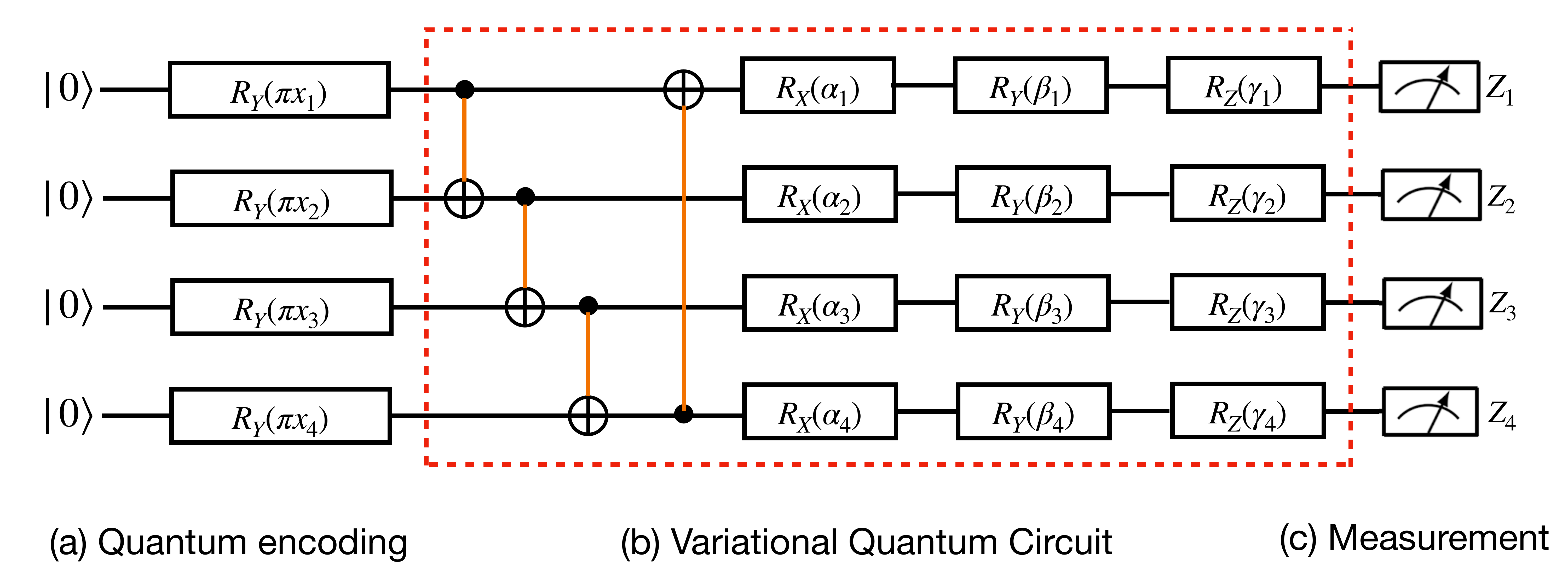, width=85mm}}
\caption{{\it A framework of quantum neural network. $R_{X}(\cdot)$, $R_{Y}(\cdot)$, $R_{Z}(\cdot)$ separately denote Pauli rotation X, Y, Z gates. The circuits in the dash square correspond to the learnable model with the repeated copies. }}
\label{fig:fig1}
\end{figure}

\begin{enumerate}
\item The framework of quantum encoding bridges the relationship between the classical data input $\textbf{x}$ and its quantum state $|\textbf{x} \rangle$. In other words, quantum encoding is associated with the generation of quantum embedding from the classical input vector $\textbf{x} = [x_{1}, x_{2}, x_{3}, x_{4}]^{T}$. The quantum state $|\textbf{x} \rangle$ can be written as Eq.~(\ref{eq:xx}). 
\begin{equation}
\label{eq:xx}
\begin{split}
& |\textbf{x} \rangle = |x_{1} \rangle \otimes |x_{2} \rangle \otimes |x_{3} \rangle \otimes |x_{4} \rangle \\
&= \left[\begin{matrix} \cos(x_{1})  \\  \sin(x_{1})  \end{matrix} \right]\otimes \left[\begin{matrix} \cos(x_{2})  \\  \sin(x_{2})  \end{matrix} \right] \otimes \left[\begin{matrix} \cos(x_{3})  \\  \sin(x_{3})  \end{matrix} \right] \otimes \left[\begin{matrix} \cos(x_{4})  \\  \sin(x_{4})  \end{matrix} \right]  \\
&= \left(\otimes_{i=1}^{4} R_{Y}(2x_{i})\right) |0\rangle^{\otimes 4} .
% |\textbf{x} \rangle &= |x_{1} \rangle \otimes |x_{2} \rangle \otimes |x_{3} \rangle \otimes |x_{4} \rangle  \\
 %			    &=\left[\begin{matrix} \cos(\pi x_{1})  \\  \sin(\pi x_{1})  \end{matrix} \right]\otimes \left[\begin{matrix} \cos(\pi x_{2})  \\  \sin(\pi x_{2})  \end{matrix} \right] \otimes \left[\begin{matrix} \cos(\pi x_{3})  \\  \sin(\pi x_{3})  \end{matrix} \right] \otimes \left[\begin{matrix} \cos(\pi x_{4})  \\  \sin(\pi x_{4})  \end{matrix} \right],
% 			   &=\left[\begin{matrix} \cos(x_{1})  \\  \sin(x_{1})  \end{matrix} \right]\otimes \left[\begin{matrix} \cos(x_{2})  \\  \sin(x_{2})  \end{matrix} \right] \otimes \left[\begin{matrix} \cos(x_{3})  \\  \sin(x_{3})  \end{matrix} \right] \otimes \left[\begin{matrix} \cos(x_{4})  \\  \sin(x_{4})  \end{matrix} \right],
 \end{split}
\end{equation}
Then, the circuits of quantum encoding in Figure~\ref{fig:fig1} produce the following quantum state as Eq. (\ref{eq:3}).
\begin{equation}
\label{eq:3}
\begin{split}
&\hspace{2mm} \left( \otimes_{i=1}^{4} R_{Y}(\pi x_{i}) \right) |0 \rangle^{\otimes 4} \\
&= \left[\begin{matrix} \cos(\pi x_{1})  \\  \sin( \pi x_{1})  \end{matrix} \right]\otimes \left[\begin{matrix} \cos( \pi x_{2})  \\  \sin(\pi x_{2})  \end{matrix} \right] \otimes \left[\begin{matrix} \cos(\pi x_{3})  \\  \sin(\pi x_{3})  \end{matrix} \right] \otimes \left[\begin{matrix} \cos(\pi x_{4})  \\  \sin(\pi x_{4})  \end{matrix} \right] .
\end{split}
\end{equation}

\item The model of the VQC in the dashed square consists of CNOT gates and adjustable rotation gates $R_{X}, R_{Y}, R_{Z}$. The CNOT gates mutually impose quantum entanglement between any two quantum wires, so that the qubits from all the wires can be entangled. The rotation angles $\alpha_{i}$, $\beta_{i}$, and $\gamma_{i}$ are adjustable and can be taken as the trainable parameters for $R_{X}$, $R_{Y}$, and $R_{Z}$, respectively. 

%The rotation gates $R_{X}$, $R_{Y}$ and $R_{Z}$, which are associated with the unitary matrices in Figure~\ref{fig:fig2}, stand for a linear mapping between quantum inputs and quantum outputs. 

\item The outputs of the quantum states should be projected by measuring the expectation values of $4$ observables $\hat{\textbf{z}} = [\hat{z}_{1}, \hat{z}_{2}, \hat{z}_{3}, \hat{z}_{4}]$ for a classical vector $\textbf{z}$ as Eq.~(\ref{eq:4}).
\begin{equation}
\label{eq:4}
\mathcal{M}: |x \rangle \rightarrow \textbf{z} = \langle x| \hat{\textbf{z}} | x \rangle. 
\end{equation}

\end{enumerate}

\begin{figure}[ht]
\centerline{\epsfig{figure=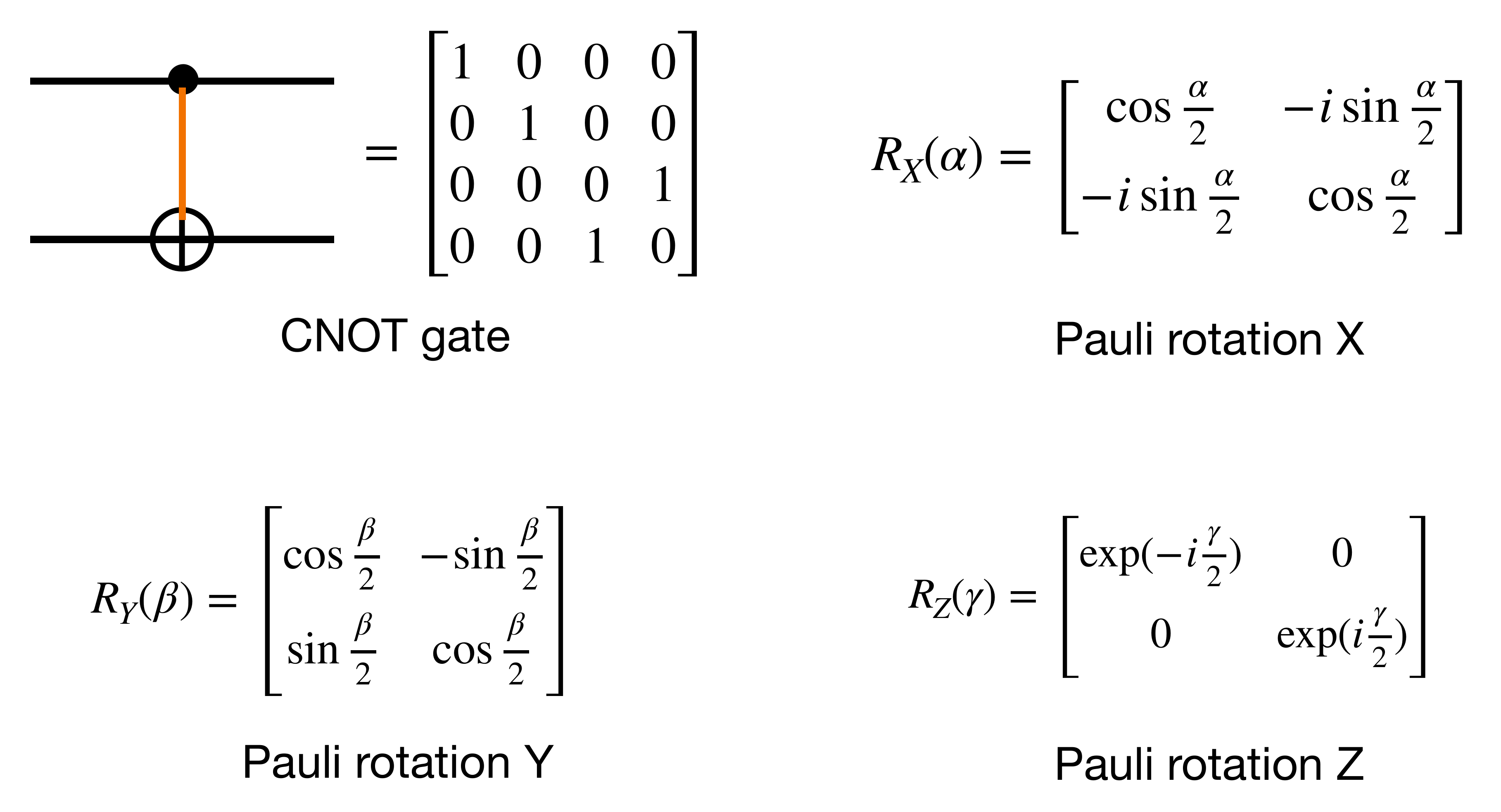, width=85mm}}
\caption{{\it Unitary matrices of the quantum gates. }}
\label{fig:fig2}
\end{figure}

%Moreover, Figure~\ref{fig:fig2} shows the unitary matrices associated with the quantum gates (CNOT gate, and Pauli rotation X, Y, Z gates~\cite{kaye2007introduction}) as shown in Figure~\ref{fig:fig1}. Based on the mathematical expression in Figure~\ref{fig:fig2}, the framework of the QNN in Figure~\ref{fig:fig1} corresponds to a linear operator because all the quantum gates are associated with unitary linear matrices. 
Moreover, the rotation gates $R_{X}$, $R_{Y}$, and $R_{Z}$, which are associated with the unitary matrices in Figure~\ref{fig:fig2}, stand for a linear mapping between quantum inputs and quantum outputs. One key advantage of the QNN is that fewer model parameters are involved in the QNN model. For example, Figure~\ref{fig:fig1} shows 4 quantum wires and 6 VQC layers, which result in 72 trainable model parameters in the QNN. 

Besides, the QNN model can be trained in an end-to-end pipeline based on the back-propagation algorithm with different stochastic gradient descent (SGD) optimizers such as Adam~\cite{kingma2014adam}, RMSprop~\cite{zou2019sufficient}, and Adadelta~\cite{zeiler2012adadelta}. This is because the update of the QNN parameters follows a first-order optimization technique to minimize a loss function over the dataset, which is represented as Eq.~(\ref{eq:grad}). 
\begin{equation}
\label{eq:grad}
\Theta \leftarrow \Theta - \eta\frac{\partial \mathcal{L}(\Theta)}{\partial \Theta}, 
\end{equation}
where $\Theta = [\Theta_{1}, \Theta_{2}, ..., \Theta_{d}]^{T}$ are the parameters to be learnt, $\mathcal{L}$ is the loss over the data, and $\eta$ is the learning rate. Given a small $\epsilon$, the partial derivative term can be approximated by using the finite difference method as Eq.~(\ref{eq:6}). 
\begin{equation}
\label{eq:6}
\frac{\partial C(\Theta_{i})}{\partial \Theta_{i}} = \frac{\mathcal{L}(\Theta_{i} + \epsilon) - \mathcal{L}(\Theta_{i} - \epsilon) }{2\epsilon} + \mathcal{O} (\epsilon^{2}). 
\end{equation}

\section{Hybrid Classical-Quantum QNN for SCR}
\label{sec4}
\begin{figure}[ht]
\centerline{\epsfig{figure=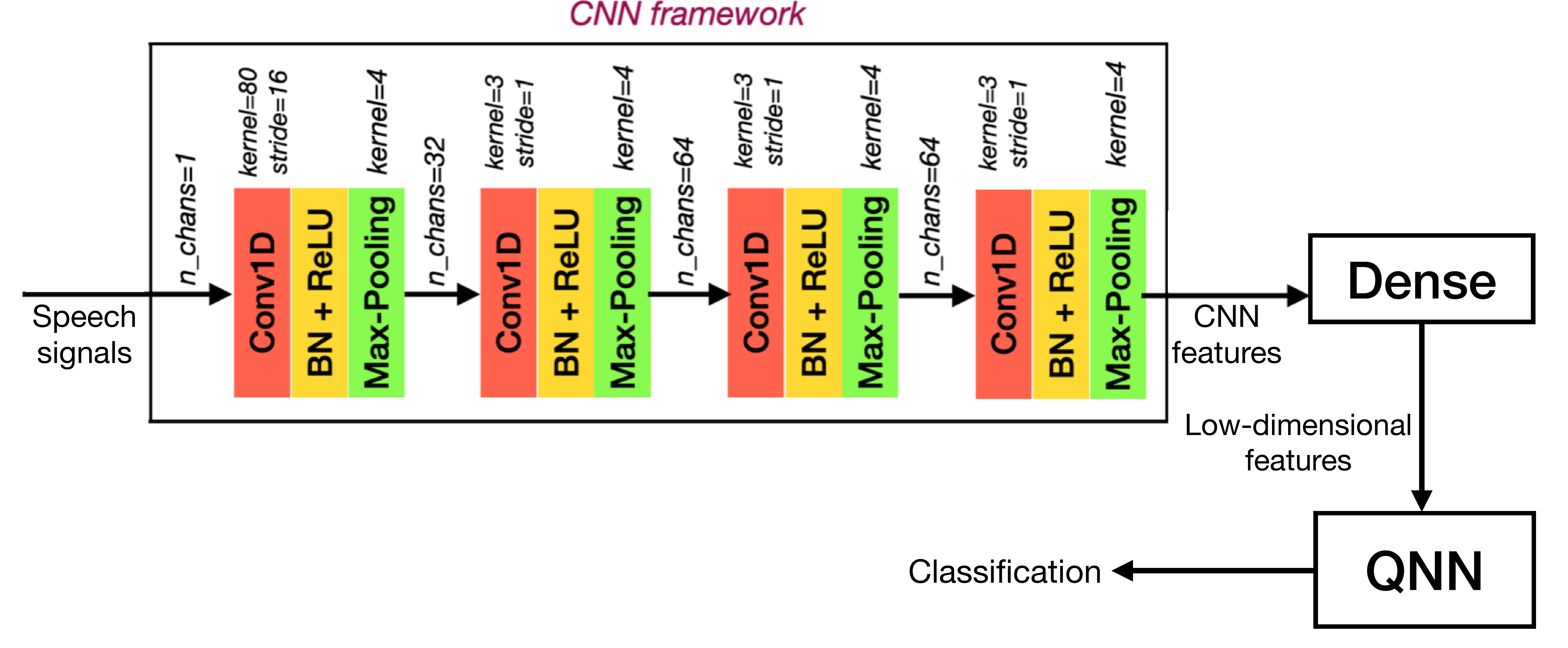, width=90mm}}
\caption{{\it A classical-quantum hybrid QNN for SCR. CNN is utilized as feature extraction and QNN is applied for classification. }}
\label{fig:fig3}
\end{figure}

Next, we illustrate the architecture of the hybrid classical-quantum QNN model for SCR in Figure~\ref{fig:fig3}. The CNN framework consists of four 1D convolutional layers (Conv1D) followed by batch normalization (BN) and the ReLU activation. A max-pooling layer with a kernel of $4$ is also used after each Conv1D. The output of the CNN framework is a set of high-dimensional CNN features, which should be compressed to low-dimensional features (Dense) connected to the quantum encoding framework in a QNN. The encoded quantum states go through the QNN model and the measured outputs correspond to the classification labels for a certain task. The output of the QNN is connected to a classification layer by a non-trainable matrix. This hybrid classical-quantum QNN is denoted as CNN-QNN. 

\begin{figure}[ht]
\centerline{\epsfig{figure=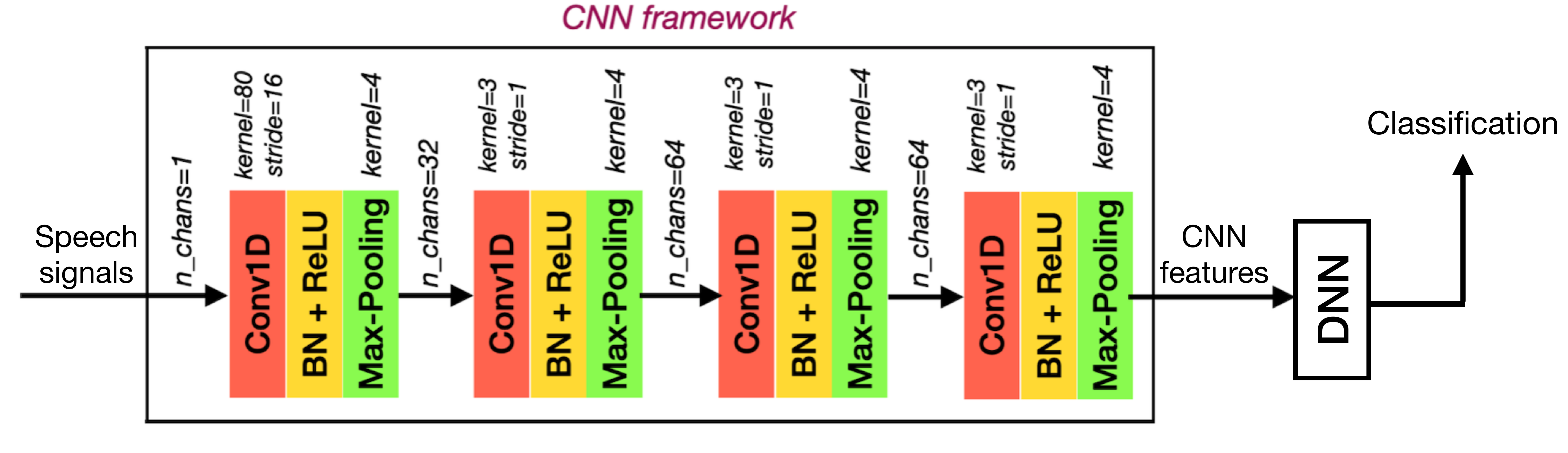, width=95mm}}
\caption{{\it A classical CNN-DNN for SCR. CNN is utilized for feature extraction and DNN is used for classification.}}
\label{fig:fig4}
\end{figure}

Accordingly, a CNN-DNN model is shown in Figure~\ref{fig:fig4}, where the QNN is replaced by a classical DNN. The feature reduction component is removed, but more model parameters are included in the DNN model. 

\section{Classical-to-Quantum Transfer Learning }
\label{sec5}
As discussed in the introduction, hybrid classical-to-quantum transfer learning is significantly appealing in the current technological era of NISQ devices. Although there are so many technical limitations in the usage of quantum computing nowadays, e.g., a small number of available qubits and noisy quantum circuits, the NISQ computers are approaching the quantum supremacy milestone~\cite{arute2019quantum, harrow2017quantum}. At the same time, we could make use of the very successful and well-tested tools of classical deep learning, especially for speech and language processing tasks in which some specific datasets are not large enough to attain well-trained neural networks. However, prior knowledge of some networks that are pre-trained on generic datasets can be shared with a new model for a specific task.

\begin{figure}[ht]
\centerline{\epsfig{figure=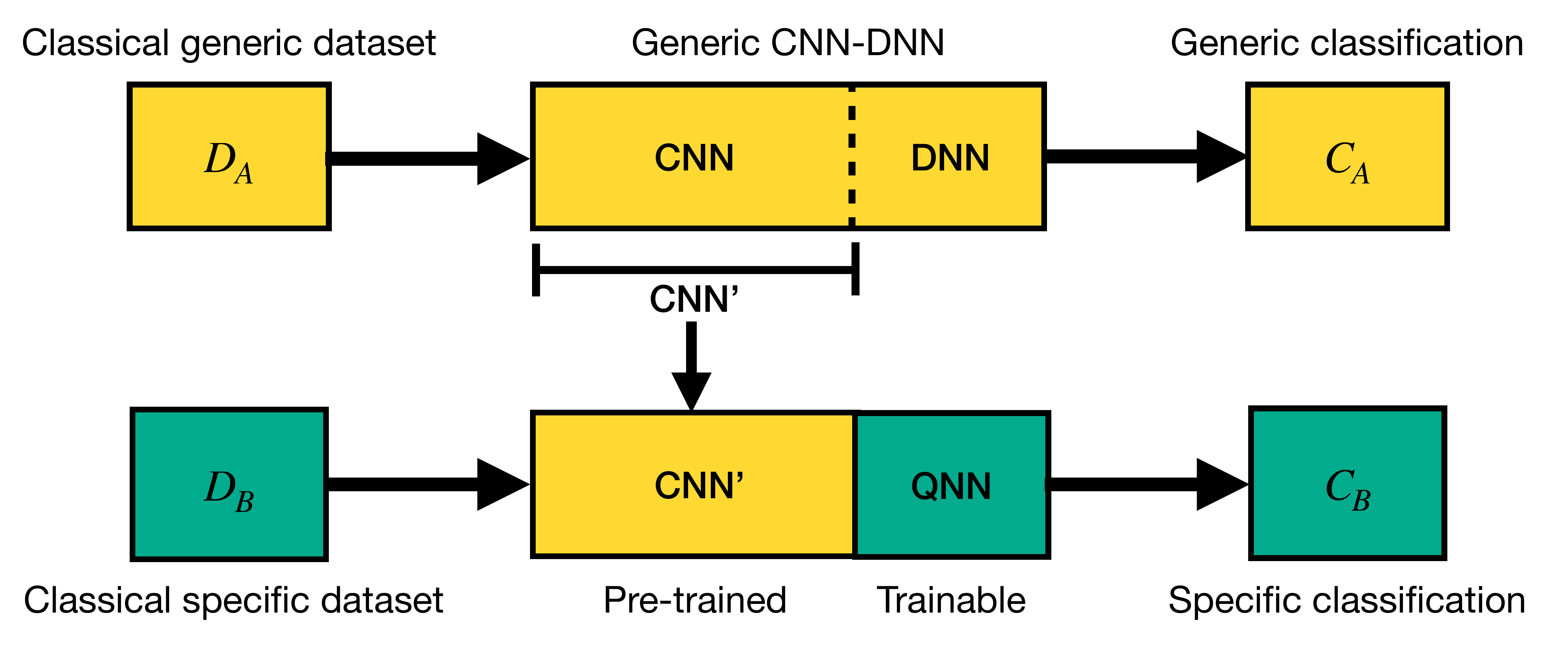, width=85mm}}
\caption{{\it An illustration of hybrid transfer learning. A generic CNN-DNN is pre-trained, and the CNN model is transferred to the CNN-QNN. The parameters of the QNN need further fine-tuning based on a classical specific dataset.}}
\label{fig:fig5}
\end{figure}

In classical machine learning, transfer learning has been widely used in certain applications. For example, the generative pre-trained transformer (GPT)~\cite{brown2020language} and bidirectional encoder representations from transformers (BERT)~\cite{devlin2018bert} are always adapted to particular language processing tasks by simply appending a few neural networks on top of the pre-trained language models. Similarly, a hybrid classical-to-quantum transfer learning algorithm comprises of applying exactly those classical pre-trained models as feature extractors and then post-processing these features on a quantum computer. As for our transfer learning algorithm applied to SCR, a pre-trained generic CNN-DNN model is obtained beforehand based on a classical generic dataset as shown in Figure~\ref{fig:fig5}. The CNN model of the pre-trained CNN-DNN, namely CNN', is transferred to the CNN-QNN, which generates CNN'-QNN. During the training stage of the CNN'-QNN, the CNN' model is fixed, and only the parameters of the QNN need further fine-tuning on the classical specific dataset. Since the transferred CNN is not involved in the training process, a small number of specific training data is enough to optimize the model parameters of the QNN. Besides, the hybrid classical-to-quantum transfer learning significantly speeds up the training efficiency because the overhead of the frequent communication between classical and quantum devices can be avoided. 

\section{Experiments}
\label{sec6}
In this section, we assess our hybrid transfer learning algorithm for CNN-QNN in the following aspects:
\begin{enumerate}
%\item We train a CNN-QNN in an end-to-end learning paradigm, where parameters from both CNN and QNN are jointly optimized.
%\item The CNN model from a pre-trained CNN-DNN is transferred to the CNN part of the CNN-DNN and does not update model parameters.
%\item The pre-trained CNN in CNN-DNN is transferred to the CNN-QNN, and the parameters from QNN and CNN are jointly updated.
\item Performance gain can be obtained by applying the hybrid transfer learning technique. 
\item The hybrid transfer learning can work with both noiseless and noisy quantum circuits.
\end{enumerate}

\subsection{Data profile}

Our SCR task was conducted on the Google speech command dataset~\cite{warden2018speech}. The dataset is composed of $11,165$ development and $6,500$ test utterances, which come from $35$ spoken commands, e.g., [`left', `go', `yes', `down', `up', `on', `right', ...]. The development data are randomly split into two parts: $90\%$ is used for model training and $10\%$ is used for model validation. All the audio files are about $1$ second long, downsampled from $16$KHz to $8$KHz. The batch size was set to 256 in the training process, and the speech signals in a batch were configured as the same length by zero padding. 

\subsection{Experimental setup}
Our baseline CNN-DNN system is based on the classical neural network architecture as shown in Figure~\ref{fig:fig4}, and our proposed CNN-QNN model is shown in Figure~\ref{fig:fig3}. The two models share the same CNN framework, which consists of 4 CNN blocks. Each CNN block is composed of a Conv1D followed by BN and ReLU, and a max-pooling is used to reduce the dimension of the speech features. The kernel size and stride are set as $80$ and $16$, for the first CNN block, and $3$ and $1$, respectively for the other CNN blocks. Moreover, the number of channels for each block follows the $1-32-64-64$ configuration. Besides, the kernel size was configured as $4$ for the max-pooling layer of each CNN block. The time-series signal is directly fed into the first CNN block, and the output of the CNN framework is related to $64$-dimensional abstract features. 

As for the classical CNN-DNN, the setup of the DNN is as follows: the hidden layers of the DNN were configured as $64-128-256-512$, and the ReLU activation function was imposed upon each hidden layer except the top hidden layer. The output of the CNN-DNN was connected to $35$ classes, which correspond to the spoken commands.  

For the setup of the CNN-QNN, the number of qubits was set as $8$ and the classical abstract features need to be further compressed to $8$ dimensions. The $8$ classical features were encoded into quantum embeddings that go through the VQC model. Since 8 channel wires were applied and $4$ repetitive VQC models were used, this leads to $96$ adjustable parameters in total. 

\subsection{Experimental results with noiseless quantum circuits}

We first examine the performance of the CNN-QNN end-to-end model, where the noiseless quantum circuits are considered. In addition to the comparison with the CNN-DNN model, our method is also compared with prominent neural network models available in the literature, namely: DenseNet-121 benchmark~\cite{mcmahan2018listening}, Attention-RNN~\cite{de2018neural, yang2020characterizing}, and QCNN~\cite{yang2021decentralizing}. QCNN denotes the use of quantum convolutional features for the task. We extend the $10$-class training setup in ~\cite{yang2021decentralizing} to 35 classes. Moreover, all deployed models are trained with the same SCR dataset from scratch, without any data augmentation or pre-training techniques to make a fair architecture-wise study.  

\begin{table}[tpbh]\footnotesize
\center
\renewcommand{\arraystretch}{1.3}
\begin{tabular}{|c||c|c|c|c|}
\hline
Models      		& Params.  $(\text{Mb})$		&   CE     	& Acc.  ($\%$)   \\
\hline
DenseNet-121~\cite{mcmahan2018listening}				&   7.978		&   0.473 	 	&    82.11			\\
\hline
Attention-RNN~\cite{de2018neural}				&   0.170		&   0.291 	 	&    93.90 			\\
\hline
QCNN~\cite{yang2021decentralizing}				&   0.186		&   0.280 	 	&    94.23			\\
\hline
CNN-DNN			&   0.216		&   0.251 	 	&    94.42 			\\
\hline
CNN-QNN			&    $\textbf{0.071}$		&   0.437		&   83.25			\\
\hline
\end{tabular}
\caption{The experimental results on the test dataset. Params. represents the number of model parameters; CE means the cross-entropy; and Acc. refers to the classification accuracy.}
\label{tab:tab6}
\end{table}

Table~\ref{tab:tab6} shows the empirical results of various models. Although the classical CNN-DNN outperforms the DenseNet-121, Attention-RNN, and QCNN systems in terms of lower CE value and higher accuracy, our CNN-QNN model with much fewer parameters cannot reach improvement. This motivates us to investigate the hybrid classical-to-quantum transfer learning. 

CNN-QNN$\_$2 denotes a CNN-QNN model in which the CNN model is transferred and fixed. In other words, only the VQC parameters are trainable in the CNN-QNN$\_2$ model. By comparison, CNN-QNN$\_3$ denotes that the CNN comes from a pre-trained CNN-DNN, and both CNN and QNN parameters need to be updated during the training stage. 
\begin{table}[tpbh]\footnotesize
\center
\renewcommand{\arraystretch}{1.3}
\begin{tabular}{|c||c|c|c|c|}
\hline
Models      & Params. $(\text{Mb})$	&   CE     	&    Acc.  ($\%$)     			 \\
\hline
CNN-DNN		&  0.216		&   0.251 	 	&    94.42 			\\
\hline
CNN-QNN$\_$2	&   \textbf{0.00096}			& $ \textbf{0.248}$	 &   94.58			 \\
\hline
CNN-QNN$\_$3	& $0.071$	& 	0.267	       &  	\textbf{94.87}			 \\
\hline
\end{tabular}
\caption{The experimental results on the test dataset. The experiments were conducted with noiseless quantum circuits. }
\label{tab:tab7}
\end{table}

The results of the hybrid transfer learning are shown in Table~\ref{tab:tab7}. Compared with CNN-DNN, CNN-QNN$\_2$ attains better accuracy ($94.58\%$ vs. $94.42\%$) with the lowest CE value (0.248 vs. 0.251) and much fewer parameters ($0.00096$ vs. $0.216$). On the other hand, CNN-QNN$\_3$ achieves the best accuracy performance. The simulation results on noiseless circuits suggest the effectiveness of hybrid transfer learning. 

\subsection{Experimental results with noisy quantum circuits}
Next, we discuss the hybrid transfer learning algorithm on the NISQ device, where noisy quantum circuits are considered. More specifically, we follow an established noisy circuit experiment with the NISQ device suggested by~\cite{chen2020variational}. One major advantage of the setups is to observe the robustness and preserve the quantum advantages of a deployed QNN with a physical setting being close to quantum processing unit (QPU) experiments. 

As for the detailed setup, we first use an IBM Q 20-qubit machine to collect channel noise in the real scenario for a deployed QNN and then upload the machine noise into our Pennylane-Qiskit~\cite{bergholm2018pennylane} simulator. As shown in Table~\ref{tab:tab8}, the hybrid transfer learning algorithm still achieves good results with much fewer parameters. Although with the noisy quantum circuits, both CNN-QNN$\_2$ and CNN-QNN$\_3$ cannot outperform the classical CNN-DNN, their results are very close. Furthermore, the empirical performance of CNN-QNN$\_$2 and CNN-QNN$\_$3 would become much better when fault-tolerant quantum computers could become available. 

\begin{table}[tpbh]\footnotesize
\center
\renewcommand{\arraystretch}{1.3}
\begin{tabular}{|c||c|c|c|c|}
\hline
Models      & Params. $(\text{Mb})$	&   CE     	&    Acc.  ($\%$)     			 \\
\hline
CNN-DNN		&  0.216		&   0.251 	 		&    94.42 			\\
\hline
CNN-QNN$\_$2	&   \textbf{0.00096}		& $0.274$	 &  93.38		 \\
\hline
CNN-QNN$\_$3	& $0.071$	& 	0.269	        &    93.84			 \\
\hline
\end{tabular}
\caption{The experimental results on the test dataset. The simulations were conducted with noisy quantum circuits.}
\label{tab:tab8}
\end{table}

\section{Conclusions}
\label{sec7}
This work focuses on a hybrid classical-to-quantum transfer learning algorithm for QNNs applied to SCR. We first set up the VQC-based QNN, and then design a CNN-QNN-based SCR system. We employ hybrid transfer learning to transfer a pre-trained CNN framework to our CNN-QNN system so that better performance could be obtained. Our experiments on the Google speech command dataset show that the hybrid classical-to-quantum transfer learning is of significance in enhancing classification accuracy and lowering cross-entropy loss value for the CNN-QNN model.

\clearpage
% References should be produced using the bibtex program from suitable
% BiBTeX files (here: strings, refs, manuals). The IEEEbib.bst bibliography
% style file from IEEE produces unsorted bibliography list.
% -------------------------------------------------------------------------
\bibliographystyle{IEEEbib}
\bibliography{refs}

\end{document}